\documentclass[11pt]{article}

\usepackage{acl}
\usepackage{enumitem}
\usepackage{times}
\usepackage{latexsym}
\usepackage{booktabs} 
\usepackage{color}    
\usepackage{amsmath}  
\usepackage{amssymb}  
\usepackage{amsfonts} 
\usepackage{algorithm}
\usepackage{algorithmic}
\usepackage[table]{xcolor}
\usepackage{graphicx}
\usepackage{lipsum}
\usepackage{stfloats}
\usepackage{hyperref}
\usepackage[T1]{fontenc}
\usepackage{stfloats}

\usepackage[utf8]{inputenc}

\usepackage{microtype}
\usepackage{inconsolata}

\usepackage{graphicx}

\title{When Attention Collapses: Stage-Aware \\Visual Token Pruning from Structure to Semantics}

\author{
\textbf{Jiahui Wang}\textsuperscript{1}\thanks{Equal contribution.}
\quad
\textbf{Kai Zhang}\textsuperscript{1}\footnotemark[1]
\quad
\textbf{Mai Han}\textsuperscript{2}
\quad
\textbf{Huanghe Zhang}\textsuperscript{1}\thanks{Corresponding author.}
\\
\textsuperscript{1}Shandong University
\\
\textsuperscript{2}National University of Singapore (Suzhou) Research Institute
\\
\texttt{wangjiahui27@mail.sdu.edu.cn},
\texttt{zhanghuanghe@sdu.edu.cn}
}

\begin{document}
\maketitle
\begin{abstract}
Vision-Language Models (VLMs) have demonstrated remarkable capabilities but suffer from significant computational overhead during inference. While visual token pruning offers a promising solution, existing methods predominantly rely on initial attention scores. This single-metric paradigm presents a critical flaw: high attention scores inherently collapse onto semantically similar regions, thereby severely reducing feature diversity and discarding vital contextual details. To address this, we introduce Structure-to-Semantics (STS), a novel two-stage visual token pruning framework that explicitly decouples the pruning process. The first stage employs a repulsion-based sampling mechanism to maximize spatial and structural diversity. The second stage leverages instruction-aware cross-attention to precisely filter out prompt-irrelevant tokens. This two-stage synergy constitutes the core of STS, first ensuring geometric coverage and then refining the retained tokens according to semantic relevance. Extensive evaluations demonstrate that STS mitigates the redundancy caused by attention-based selection, improving both structural diversity and fine-grained task alignment of the preserved visual tokens. 
\end{abstract}

\section{Introduction}

Vision-Language Models (VLMs) \citep{NEURIPS2023_6dcf277e,bai2023qwentechnicalreport,team2023gemini} have achieved strong performance across a wide range of multimodal tasks by coupling high-resolution vision encoders with powerful Large Language Models (LLMs) \citep{brown2020language,alayrac2022flamingo,radford2019language}. While this design enables fine-grained visual understanding, it also introduces substantial computational overhead. To preserve visual fidelity, modern vision encoders often produce hundreds of visual tokens per image, all of which must be processed by the LLM \citep{xu2024llavauhdlmmperceivingaspect,chen2024internvlscalingvisionfoundation}. Due to the quadratic complexity of  Transformer self-attention with respect to sequence length \citep{tay2022efficienttransformerssurvey,wen2025stoplookingimportanttokens}, this large token count leads to significant increases in inference latency and memory usage, limiting the practicality of VLMs in real-time and resource-constrained settings. 

To alleviate this computational burden, visual token pruning
\citep{zhang2025textvisualattentionexploitingvisual,zhang2025sparsevlmvisualtokensparsification,yang2026visionziplongerbetternecessary,11247780}
has emerged as a promising direction for reducing spatial and semantic redundancy.
Existing methods often rely on attention scores as token-importance indicators
\citep{chen2024imageworth12tokens,xing2025pyramiddropacceleratinglargevisionlanguage}.
However, the representation dynamics of vision encoders in VLM pipelines remain less explored compared with the attention dynamics of LLMs
\citep{xiao2024efficient,zhang2023h2oheavyhitteroracleefficient}.

By analyzing these dynamics, we identify a key limitation of attention-based visual token pruning:
high-attention tokens tend to concentrate in semantically similar regions, leading to redundant selections.
Specifically, in shallow vision layers, attention scores are only weakly related to feature similarity.
As depth increases, tokens that are close in the feature space increasingly receive similar attention scores.
As a result, pruning based solely on attention is prone to retaining multiple tokens from the same semantic neighborhood while discarding more diverse visual details.
These observations suggest that a single static pruning criterion is insufficient, motivating our decoupled, stage-aware pruning strategy.

To overcome this "clustering trap" and prevent the loss of feature diversity, we reformulate the initial token selection—prior to the LLM stage—as a potential-energy-inspired objective. Drawing inspiration from electrostatic repulsion principles, we model visual tokens as mutually repulsive charged particles in the feature space \citep{wang2020understanding,Kulesza_2012}. In this analogy, tokens with high semantic similarity generate strong repulsive interactions. This mechanism actively discourages excessive concentration in redundant regions, thereby ensuring the retention of a globally diverse and structurally complete set of visual features.

To complement this diversity-preserving mechanism, we further introduce pruning within an intermediate layer of the language model to remove task-irrelevant tokens. This two-stage design jointly preserves structural diversity and semantic relevance, yielding a pruning strategy that better matches the stage-dependent representation dynamics of VLMs. Extensive experiments across multiple vision–language models demonstrate that our method improves inference efficiency while maintaining strong task performance. 

In summary, our contributions are three-fold:
\begin{itemize}[noitemsep, topsep=0pt, leftmargin=*, labelsep=0.5em]
    \item In vision encoders, we find that tokens with high attention scores tend to concentrate in semantically similar regions, and we further quantify this phenomenon using a KNN-based analysis. This helps explain why traditional attention-based pruning methods often select spatially clustered tokens.
    \item We propose STS, a training-free, stage-aware visual token pruning framework that uses a potential-energy-inspired objective to preserve global structural diversity while maintaining task-specific semantic relevance.
    \item Extensive experiments across multiple vision--language models and benchmarks demonstrate that STS achieves favorable efficiency--performance trade-offs, improving inference efficiency while preserving strong task performance even under aggressive token reduction.
\end{itemize}

\section{Related Work}

\textbf{Vision-Language Models.} Recent vision-language models (VLMs) have achieved strong multimodal reasoning performance by encoding images into dense sequences of visual tokens. Representative architectures, including LLaVA-1.5 and LLaVA-NeXT \citep{liu2024improvedbaselinesvisualinstruction,liu2024llavanext}, employ high-resolution vision encoders that generate large numbers of visual tokens, and this token burden is further amplified in video-based extensions such as Video-LLaVA \citep{lin2024videollavalearningunitedvisual}. However, this design introduces substantial inference overhead: self-attention scales quadratically with sequence length, while KV-cache memory grows linearly with the number of tokens \citep{kwon2023efficientmemorymanagementlarge,pope2023efficiently}. As a result, processing all visual tokens is often computationally prohibitive in latency- or memory-constrained settings, motivating the need for effective token sparsification strategies that reduce redundancy without sacrificing critical visual information. 

\textbf{Visual Token Reduction.} Existing approaches typically rely on heuristic criteria, such as attention-based pruning \citep{xing2025pyramiddropacceleratinglargevisionlanguage}, token merging \citep{bolya2022token,liang2022patchesneedexpeditingvision}, and diversity-driven sampling \citep{liang2023clustsegclusteringuniversalsegmentation,wen2025stoplookingimportanttokens}. However, attention-based methods may repeatedly select tokens from similar regions, while diversity-based heuristics often capture only local differences and may still miss the overall visual structure. Motivated by these limitations, we propose STS, a training-free and stage-aware pruning framework. STS first applies a potential-energy-inspired selection strategy before the LLM stage to encourage globally diverse token coverage, and then performs task-aware pruning within the language model to remove semantically irrelevant tokens. In this way, STS explicitly balances geometric diversity with semantic relevance.

\section{Empirical Analysis}

\begin{figure*}[t] 
    \centering
    \includegraphics[width=1\textwidth]{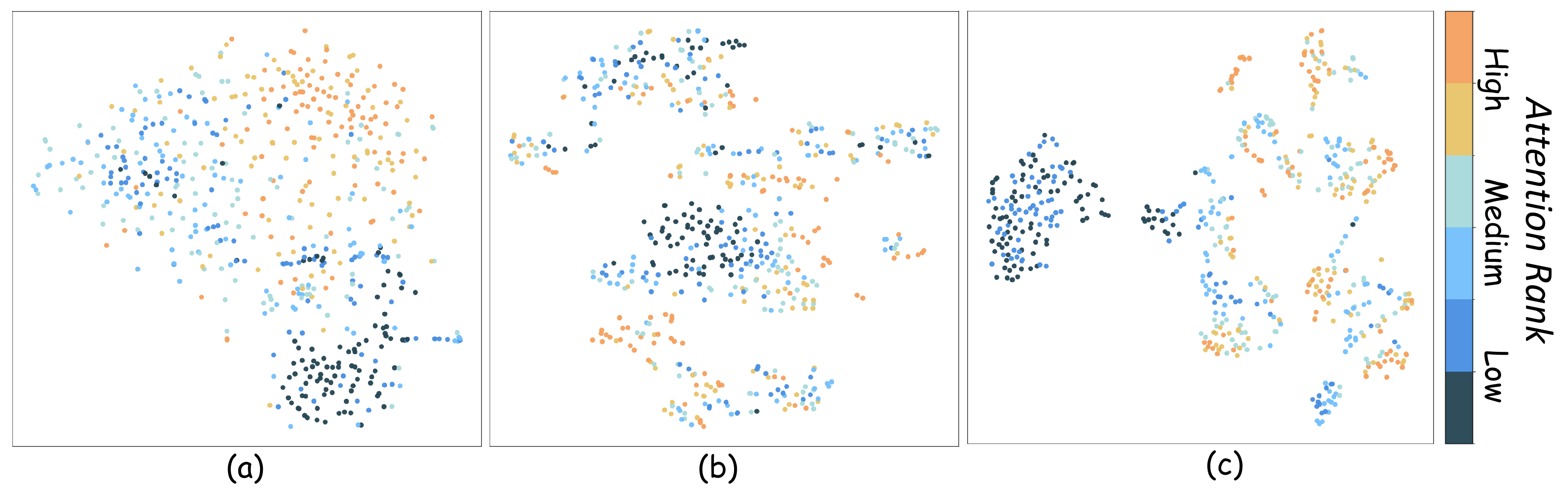} 
    
    \caption{t-SNE visualization of visual tokens for a representative image from the COCO dataset at (a) the 1st layer, (b) the 14th layer, and (c) the 24th layer. In shallow layers, attention similarity is weakly aligned with feature similarity, whereas in deeper layers, tokens with similar attention scores tend to cluster in nearby regions of the feature space. } 
    \label{fig:tsne} 
\end{figure*}

Our analysis provides two insights for visual token pruning. In the vision encoder, attention scores become increasingly similar among feature-similar tokens in deeper layers, making attention-based pruning prone to redundant selections. In the LLM decoder, visual information flow shifts from broad contextual aggregation in earlier layers to task-specific concentration in deeper layers. These findings motivate a stage-aware pruning strategy that first preserves structural diversity before the LLM and then applies semantic filtering within the LLM.

\subsection{Feature-Attention Redundancy in Vision Encoders}

To examine whether semantically similar tokens receive similar attention scores, and how this relationship evolves across the vision encoder, we introduce a KNN-based \citep{cover1967nearest} Consistency Score ($C$). This metric measures the consistency of attention scores within local neighborhoods of the feature space \citep{caron2021emergingpropertiesselfsupervisedvision}. Intuitively, if tokens that are close in the feature space also receive nearly identical attention scores, then attention-based pruning may struggle to distinguish among redundant tokens. 

Formally, let $\mathcal{T} = \{t_1, \dots, t_N\}$ denote the set of visual tokens, where each token $t_i$ is associated with a feature vector $\mathbf{f}_i$ and an attention score $s_i$. We first project all features into an $\ell_2$-normalized embedding space, and for each token $t_i$, we identify its $k$-nearest neighbors $\mathcal{N}_k(t_i)$ based on feature similarity. We then define the \emph{local fluctuation} of attention scores as
\begin{equation}
\sigma_{\text{local}} = \frac{1}{N} \sum_{i=1}^{N} \mathrm{std}\!\left(\{s_j \mid t_j \in \mathcal{N}_k(t_i)\}\right).
\end{equation}

A small $\sigma_{\text{local}}$ indicates that tokens within the same local feature neighborhood tend to receive similar attention scores. To make this quantity comparable across layers and models, we normalize it by the global standard deviation of attention scores over the full token set, denoted as $\sigma_{\text{global}}$. The resulting KNN Consistency Score is defined as
\begin{equation}
C = 1 - \frac{\sigma_{\text{local}}}{\sigma_{\text{global}}}.
\end{equation}

A larger value of $C$ indicates stronger alignment between feature similarity and attention similarity. In such cases, attention becomes less discriminative within local semantic neighborhoods, making attention-based pruning more likely to retain spatially clustered and semantically redundant tokens rather than a diverse subset of visual features.
\begin{figure}[!t] 
    \centering
    \includegraphics[width=\linewidth]{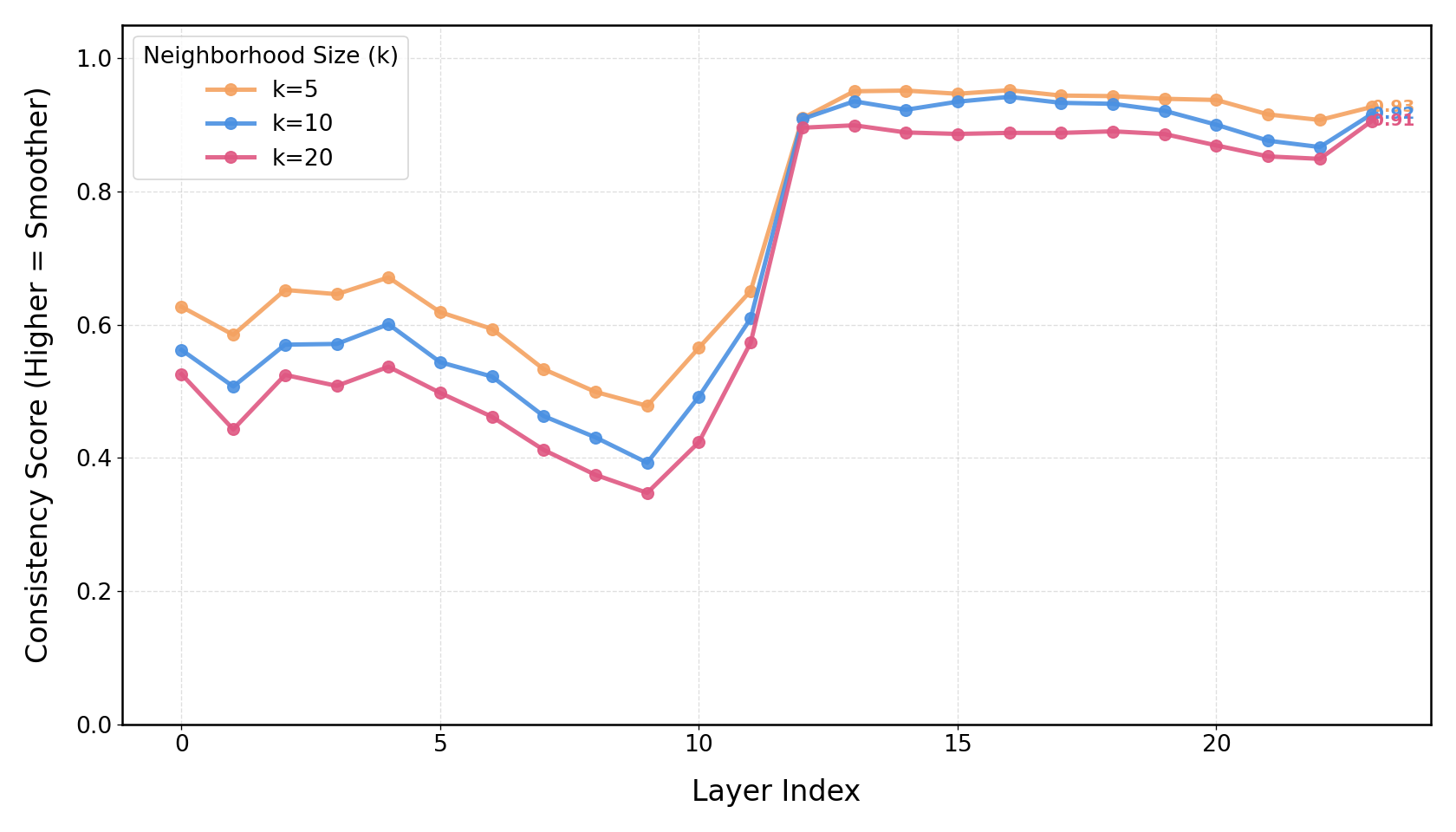} 
    \caption{KNN sensitivity analysis across LLaVA vision encoder layers. The consistency scores across various k reveal a clear shift toward feature aggregation in deeper layers, indicating an increasing alignment between attention scores and token embeddings that leads to redundancy. } 
    \label{fig:knn}
\end{figure}

\textbf{Observations.}Our analysis on LLaVA-1.5 reveals a clear depth-dependent trend. In shallow layers, the consistency score $C$ remains low, indicating that attention similarity is only weakly aligned with feature similarity. In deeper layers, however, $C$ increases markedly, showing that tokens within the same local feature neighborhood tend to receive increasingly similar attention scores. Consequently, attention-based pruning becomes prone to selecting multiple tokens from semantically similar regions, producing clustered and redundant token subsets. Such clustering can reduce the diversity of the retained tokens and increase the risk of discarding complementary long-tail visual details. These results provide a direct explanation for the failure mode of attention-based pruning in deep vision layers.

\subsection{Stage-dependent Redundancy of Visual Tokens in LLMs }

\begin{figure*}[t] 
    \centering
    \includegraphics[width=1\textwidth]{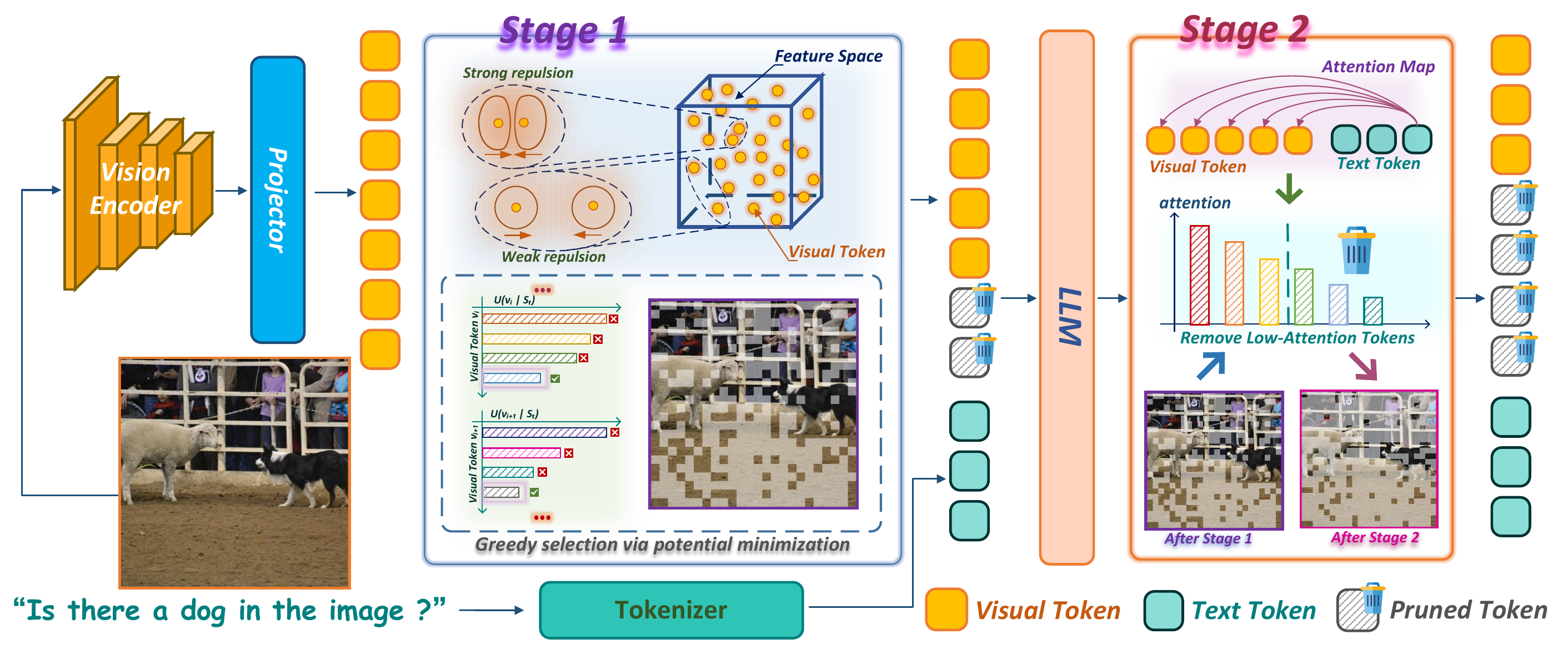} 
    
    \caption{Overview of the proposed STS framework for stage-aware visual token pruning.
Given visual tokens from the vision encoder and textual tokens from the input prompt,
STS performs token reduction in two stages. In Stage 1, visual tokens are modeled as
repulsive particles in feature space and selected using a potential-energy-inspired strategy
to promote broad coverage of the feature space. In Stage 2, the remaining tokens are further
filtered using cross-modal attention from textual tokens to visual tokens, retaining those
most relevant to the textual query. This two-stage strategy preserves both structural
diversity and task-specific semantic information while substantially reducing the number
of visual tokens processed by the LLM. } 
    \label{fig:sts} 
\end{figure*}

Building on prior analyses of attention propagation and attribution in Transformer models \citep{abnar2020quantifyingattentionflowtransformers,elhage2021mathematical}, we draw on existing findings to discuss how visual tokens are utilized across LLM layers. Rather than characterizing these layer-wise patterns in detail, we use them to motivate when visual token redundancy is likely to emerge and where pruning can be applied most effectively.

In earlier layers, the model performs broad contextual integration, distributing attention over a large set of visual tokens to capture global scene semantics. However, as computation progresses deeper into the LLM, visual processing undergoes a critical transition: guided by the textual prompt, attention gradually concentrates on a narrower, task-relevant subset of visual tokens, leaving the majority of visual features with diminishing contributions \citep{voita2019analyzingmultiheadselfattentionspecialized,zhang2024redundancyrelevanceinformationflow,dong2023attentionneedpureattention}. 

This progressive, instruction-driven concentration implies a strategic window for token reduction. Premature pruning risks discarding essential features before full contextual integration, whereas delayed pruning incurs severe computational waste by propagating redundant tokens through the deep LLM layers. Consequently, the intermediate layers emerge as the optimal pruning bottleneck—a "sweet spot" where broad context has been assimilated, yet prompt-irrelevant visual tokens can be aggressively filtered out to maximize efficiency \citep{liu2023dejavucontextualsparsity,men2024shortgptlayerslargelanguage}. Our empirical results corroborate this analysis, demonstrating that executing pruning within these early-to-middle layers successfully achieves an optimal trade-off between task accuracy and inference efficiency. We provide a detailed ablation study validating this optimal layer selection in Appendix \ref{B}. 

\section{Method}

In this section, we introduce STS, a training-free visual token pruning framework. The key idea of STS is to separate structural preservation from semantic filtering. Before visual tokens enter the LLM, textual instructions are unavailable, making it difficult to determine task relevance. Therefore, STS first selects a compact and diverse subset of visual tokens from the vision encoder output using a potential-energy-inspired objective, which promotes broad coverage of the feature space. Once the selected visual tokens are processed within the LLM, cross-modal attention provides an instruction-aware relevance signal. STS then applies task-aware filtering at an intermediate LLM layer to remove visual tokens with low relevance to the textual instruction. This two-step design first preserves structural diversity and then refines the retained tokens according to semantic relevance, reducing visual redundancy without additional training.

\subsection{Pre-LLM Pruning via Global Potential Energy Minimization}

The first stage operates on the visual tokens produced by the vision encoder before they are fed into the LLM. Let $\mathbf{V} = [v_1, v_2, \dots, v_N] \in \mathbb{R}^{N \times C}$ denote the visual token sequence, where $N$ is the number of tokens and $C$ is the feature dimension. Since textual instructions are unavailable at this stage, token selection should preserve broad structural coverage rather than depend only on local importance scores. We therefore select a compact subset $\mathcal{S} \subset \mathcal{V}$ with $|\mathcal{S}| = K \ll N$ to reduce redundancy while maintaining diverse visual information.

To achieve this, we draw inspiration from electrostatic field theory and reformulate token selection as a potential-energy minimization problem. Visual tokens are modeled as mutually repulsive particles in feature space, where interaction strengths are governed by their pairwise proximity.

\textbf{Potential Energy Modeling.} We embed the visual tokens into a metric feature space. Formally, we define the squared Euclidean distance between tokens $v_i$ and $v_j$ as
\begin{equation}
d_{ij} = \|v_i - v_j\|_2^2 .
\end{equation}
Based on this distance metric, the cumulative repulsive potential experienced by a candidate token $v_i$ with respect to the currently selected subset $\mathcal{S}_t$ is formulated as
\begin{equation}
U(v_i \mid \mathcal{S}_t) = \sum_{v_j \in \mathcal{S}_t} \frac{1}{d_{ij} + \epsilon},
\end{equation}
where $\epsilon$ is a small constant for numerical stability. This formulation assigns a larger penalty to candidates that are close to already selected tokens, thereby discouraging redundant selections from densely populated regions of the feature space. Conversely, tokens that are farther from the current selected set receive lower potential values and are more likely to be retained.

\textbf{Iterative Selection Algorithm.} We adopt an efficient greedy strategy to approximate the global minimization of this potential. To ensure deterministic behavior and improve robustness, we initialize $\mathcal{S}_0$ with an anchor token, such as the token closest to the global mean of the visual features or the token with the highest initial saliency. At each subsequent step, we select the candidate that experiences the minimum repulsive potential with respect to the already selected tokens:
\begin{equation}
v_{\text{next}} = \arg\min_{v \in \mathcal{V} \setminus \mathcal{S}_t} U(v \mid \mathcal{S}_t).
\end{equation}
By consistently selecting tokens with lower potential, the algorithm encourages the retained tokens to be well separated in the feature space. This process selects visual tokens that are more widely distributed in the feature space, reducing redundancy before they are passed to the LLM.  

\subsection{Intra-LLM Pruning via Task-Aware Filtering}

As visual tokens propagate through the LLM, the model gradually shifts from broad contextual integration to a more selective focus on task-relevant information. At this stage, purely diversity-driven retention may become suboptimal, since some structurally distinct tokens may still be irrelevant to the textual query. To address this issue, we apply task-aware pruning at an intermediate layer $L_{\text{prune}}$ \citep{yin2022adavitadaptivetokensefficient,Wang_2021}.

Let $\mathcal{S}$ denote the visual tokens retained after the pre-LLM stage, and let $t_{\mathrm{last}}$ denote the final textual token in the input sequence \citep{vaswani2017attention}. We estimate the semantic relevance of each visual token $v \in \mathcal{S}$ using the attention weight it receives from this last textual token:
\begin{equation}
R(v) = A_{t_{\mathrm{last}},v},
\end{equation}
where $A_{t_{\mathrm{last}},v}$ is the attention weight from the last textual token to visual token $v$ at layer $L_{\text{prune}}$ \citep{tang2024questqueryawaresparsityefficient}.

The final retained token set is then obtained by selecting the top-ranked visual tokens:
\begin{equation}
\mathcal{S}_{\text{final}} =
\operatorname{TopK}_{v \in \mathcal{S}}
\left( R(v), K' \right).
\end{equation}

This filtering step removes semantically irrelevant tokens while preserving those most relevant to the current instruction, thereby refining the diverse candidate pool into a compact and task-aware visual representation. For implementation details and pseudo-code, please refer to Algorithm \ref{alg:sts} .

\textbf{Remarks on FlashAttention. }Our method can still be used with models that adopt FlashAttention, since the task-aware relevance scores are obtained by recomputing the attention of the last instruction token with a standard attention operation outside the original LLM layers.

\begin{table*}[t!]
\centering
\resizebox{\textwidth}{!}{
\begin{tabular}{l|cccccccc|c}
\toprule
\multicolumn{1}{c|}{\textbf{Method}} & \textbf{GQA} & \textbf{MMB} & \textbf{MME} & \textbf{POPE} & \textbf{SQA} & \textbf{VQA$^{\text{v2}}$} & \textbf{VQA$^{\text{Text}}$} & \textbf{VizWiz} & \textbf{Avg.} \\
\midrule
\rowcolor{gray!15} LLaVA-1.5-7B & \multicolumn{9}{c}{\textit{Upper Bound, 576 Tokens} (\textbf{100\%})} \\
\textcolor{gray}{Vanilla} & 61.9 & 64.7 & 1862 & 85.9 & 69.5 & 78.5 & 58.2 & 50.1 & 100.0\% \\
\midrule
\rowcolor{gray!15} LLaVA-1.5-7B & \multicolumn{9}{c}{\textit{Retain 128 Tokens} (\textcolor{green!50!black}{$\downarrow$ 77.8\%})} \\
VisionZip (CVPR2025)& 57.6 & 62.0 & 1762 & 83.2 & 68.9 & 75.0 & 56.8 & 49.6 & 96.5\% \\
SparseVLM (ICML2025)& 56.0 & 60.0 & 1696 & 80.5 & 67.1 & 73.8 & 54.9 & 51.0 & 94.3\% \\
DART (EMNLP2025)& 57.9 & 60.7 & 1721& 80.4 & \textbf{69.1} & 74.7 & 56.3 & 52.8 & 96.8\% \\
DivPrune (CVPR2025)& 59.3 & 61.5 & 1718 & 86.7 & 68.6 & 76.0 & 56.0 & 52.8 & 97.6\% \\
Zoo-Prune (CVPR2026)& 59.5 & 61.9 & 1751 & 87.1 & 68.9 & 76.6 & \textbf{57.9} & - & 97.6\% \\
AgilePrune (ICLR2026)& 59.4 & 61.8 & 1748 & \textbf{87.4} & 68.6 & 76.4 & 57.0 & \textbf{53.0} & 98.4\% \\
\rowcolor{orange!15} STS (Ours) & \textbf{60.1} & \textbf{63.5} & \textbf{1803} & 87.2 & 68.8 & \textbf{77.3} & 57.6 & 52.5 & \textbf{99.4\%} \\
\midrule
\rowcolor{gray!15} LLaVA-1.5-7B & \multicolumn{9}{c}{\textit{Retain 64 Tokens} (\textcolor{green!50!black}{$\downarrow$ 88.9\%})} \\
VisionZip (CVPR2025)& 55.1 & 60.1 & 1690 & 77.0 & 69.0 & 72.4 & 55.5 & 51.9 & 94.1\% \\
SparseVLM (ICML2025)& 52.7 & 56.2 & 1505 & 75.1 & 67.2 & 68.2 & 51.8 & 49.6 & 89.0\% \\
DART (EMNLP2025)& 54.7 & 59.5 & 1692& 73.8 & \textbf{69.3} & 71.3 & 54.7 & 53.5 & 93.9\% \\
DivPrune (CVPR2025)& 57.8 & 59.3 & 1674 & 85.6 & 68.2 & 74.1 & 54.7 & 53.6 & 94.8\% \\
Zoo-Prune (CVPR2026)& 58.5 & 60.2 & 1675 & 85.9 & 68.3 & 75.0 & 55.4 & - & 95.2\% \\
AgilePrune (ICLR2026)& 57.4 & 60.7 & 1703 & 84.1 & 68.6 & 75.5 & 56.0 & \textbf{54.0} & 96.9\% \\
\rowcolor{orange!15} STS (Ours) & \textbf{59.0} & \textbf{61.6} & \textbf{1718} & \textbf{87.0} & 69.2 & \textbf{75.9} & \textbf{56.8} & 53.0 & \textbf{98.0\%} \\
\midrule
\rowcolor{gray!15} LLaVA-1.5-7B & \multicolumn{9}{c}{\textit{Retain 32 Tokens} (\textcolor{green!50!black}{$\downarrow$ 94.4\%})} \\
VisionZip (CVPR2025)& 51.8 & 57.0 & 1579 & 69.4 & 69.1 & 67.1 & 53.1 & 52.4 & 89.8\% \\
DART (EMNLP2025)& 52.9 & 58.5 & 1601& 69.1 & 69.3 & 67.1 & 52.2 & 52.5 & 90.9\% \\
DivPrune (CVPR2025)& 54.9 & 57.6 & 1594 & 81.5 & 68.6 & 71.2 & 52.9 & 53.3 & 93.1\% \\
AgilePrune (ICLR2026)& 54.1 & \textbf{60.4} & 1603 & 80.1 & 69.0 & \textbf{74.0} & 54.5 & \textbf{53.4} & 94.2\% \\
\rowcolor{orange!15} STS (Ours) & \textbf{57.1} & 60.2 & \textbf{1652} & \textbf{85.4} & \textbf{69.4} & 73.9 & \textbf{55.4} & 52.7 & \textbf{96.0\%} \\
\bottomrule
\end{tabular}
}
\caption{Performance comparison of different token pruning methods on LLaVA-1.5-7B across multiple benchmarks. The orange background highlights our method.}
\label{tab:7b}
\end{table*}

\section{Experiments}

\subsection{Experimental Setup}

We evaluate STS on four representative large multimodal models (LMMs)—LLaVA-v1.5-7B/13B, LLaVA-NeXT-7B, and Qwen2.5-VL-7B—covering fixed-resolution, high-resolution, and dynamic visual encoding schemes. Experiments are conducted on eight widely adopted image-based benchmarks: GQA \citep{hudson2019gqanewdatasetrealworld}, MMBench \citep{li2023evaluatingobjecthallucinationlarge}, MME \citep{fu2025mmecomprehensiveevaluationbenchmark}, POPE \citep{pope2023efficiently}, ScienceQA \citep{lu2022learnexplainmultimodalreasoning}, TextVQA \citep{singh2019vqamodelsread}, VQA-v2 \citep{goyal2017makingvvqamatter}, and VizWiz \citep{gurari2018vizwizgrandchallengeanswering} . For all benchmarks, we follow their default settings and official evaluation metrics. 

We adopt the standard inference settings of the evaluated LVLMs and report results using the official metrics of each benchmark. STS reduces visual tokens in two stages. For the intra-LLM stage, we perform task-aware filtering at a fixed intermediate layer, setting $L_{\text{prune}} = 16$ for LLaVA-series models and $L_{\text{prune}} = 14$ for Qwen2.5-VL. By default, the intra-LLM retention ratio is fixed to $\rho_{\text{intra}} = 33.3\%$, while the pre-LLM token budget is adjusted according to the target average number of visual tokens processed by the LLM. For example, to obtain an average budget of 128 tokens, we retain 192 tokens before the LLM and further reduce them to 64 tokens after intra-LLM pruning. These settings are fixed for simplicity and to enable consistent computation and efficiency measurement across different models and experimental settings. As we show later in the ablation studies, performance is relatively insensitive to the exact pruning layer within the middle stage of the LLM. 

\subsection{Comparison on Diverse Tasks }

\textbf{Results on LLaVA-1.5-7B.} As shown in Table \ref{tab:7b}, STS consistently achieves the best average performance across different token budgets on LLaVA-1.5-7B. With 128 retained tokens, STS reaches 99.4\% relative performance, outperforming strong recent baselines such as AgilePrune and Zoo-Prune. Under more aggressive pruning, STS retains 98.0\% relative performance with only 64 tokens, while achieving the best results on GQA, MMB, MME, POPE, VQA$^{\text{v2}}$, and VQA$^{\text{Text}}$. Even at the extreme 32-token budget, STS maintains 96.0\% relative performance, surpassing AgilePrune by 1.8 points and showing strong robustness under severe visual token reduction.

\textbf{Results on LLaVA-NeXT-7B.} On the high-resolution LLaVA-NeXT-7B setting, STS remains effective across different token budgets, as shown in Table \ref{tab:next}. With 640 retained tokens, STS achieves 97.6\% relative performance, slightly outperforming VisionZip and AgilePrune. Under more aggressive pruning, STS retains 96.8\% relative performance with 320 tokens and 95.3\% with only 160 tokens. Notably, at the 160-token budget, STS outperforms Zoo-Prune by 1.8 points on average and achieves the best results on GQA, MME, POPE, and VQA$^{\text{v2}}$, indicating strong performance under high-resolution visual token compression.

We focus on the main experimental results in this section. Additional comparisons and ablation studies are included in Appendix \ref{B} due to space constraints.

\begin{table*}[t!]
\centering
\resizebox{\textwidth}{!}{
\begin{tabular}{l|cccccccc|c}
\toprule
\multicolumn{1}{c|}{\textbf{Method}} & \textbf{GQA} & \textbf{MMB} & \textbf{MME} & \textbf{POPE} & \textbf{SQA} & \textbf{VQA$^{\text{v2}}$} & \textbf{VQA$^{\text{Text}}$} & \textbf{VizWiz} & \textbf{Avg.} \\
\midrule
\rowcolor{gray!15} LLaVA-NeXT-7B & \multicolumn{9}{c}{\textit{Upper Bound} (\textbf{100\%})} \\
\textcolor{gray}{Vanilla} & 69.2 & 67.9 & 1842 & 86.4 & 70.2 & 80.1 & 61.3 & 55.2 & 100.0\% \\
\midrule
\rowcolor{gray!15} LLaVA-NeXT-7B & \multicolumn{9}{c}{\textit{Retain 640 Tokens} (\textcolor{green!50!black}{$\downarrow$ 77.8\%})} \\
SparseVLM (ICML2025)& 60.3 & 65.7 & 1772 & 85.2 & 67.7 & 77.1 & 57.8 & 53.6 & 95.3\% \\
VisionZip (CVPR2025)& 61.3 & 65.8 & 1787 & 86.3 & 68.1 & 79.1 & \textbf{60.2} & \textbf{57.1} & 97.5\% \\
DART(EMNLP2025)& 61.3 & 64.9 & 1781& 85.0 & \textbf{68.2} & 78.3 & 59.5 & 57.0 & 96.8\% \\
DivPrune (CVPR2025)& 61.6 & 65.4 & 1773 & 85.5 & 67.8 & 78.9 & 55.4 & 55.7 & 95.9\% \\
AgilePrune (ICLR2026)& 62.0 & \textbf{65.9} & - & 86.1 & 67.8 & 79.3 & 59.0 & 56.0 & 97.1\% \\
Zoo-Prune (CVPR2026)& 62.2 & 65.2 & \textbf{1816} & 86.8 & 68.0 & 79.6 & 58.0 & - & 96.6\% \\
\rowcolor{orange!15} STS (Ours) & \textbf{63.4} & \textbf{65.9} & 1801 & \textbf{87.6} & 67.7 & \textbf{79.9} & 58.9 & 55.7 & \textbf{97.6\%} \\
\midrule
\rowcolor{gray!15} LLaVA-NeXT-7B & \multicolumn{9}{c}{\textit{Retain 320 Tokens} (\textcolor{green!50!black}{$\downarrow$ 88.9\%})} \\
SparseVLM (ICML2025)& 57.7 & 64.3 & 1684 & 78.6 & 67.3 & 73.4 & 55.9 & 54.2 & 92.2\% \\
VisionZip (CVPR2025)& 59.3 & 63.1 & 1702 & 82.1 & 67.3 & 76.2 & \textbf{58.9} & 56.2 & 94.4\% \\
DART (EMNLP2025)& 59.5 & 64.2 & 1743& 81.0 & 67.5 & 75.7 & 57.6 & \textbf{56.8} & 94.5\% \\
DivPrune (CVPR2025)& 59.6 & 63.7 & 1731 & 83.5 & \textbf{67.8} & 76.6 & 53.9 & 55.6 & 93.9\% \\
AgilePrune (ICLR2026)& 60.1 & 64.5 & - & 84.0 & 67.3 & 77.8 & \textbf{58.9} & 55.8 & 95.6\% \\
Zoo-Prune (CVPR2026)& 60.9 & \textbf{64.9} & 1787 & 85.5 & \textbf{67.8} & 78.1 & 57.3 & - & 95.3\% \\
\rowcolor{orange!15} STS (Ours) & \textbf{61.5} & 64.5 & \textbf{1791} & \textbf{87.5} & 67.7 & \textbf{78.8} & 58.6 & 56.3 & \textbf{96.8\%} \\
\midrule
\rowcolor{gray!15} LLaVA-NeXT-7B & \multicolumn{9}{c}{\textit{Retain 160 Tokens} (\textcolor{green!50!black}{$\downarrow$ 94.4\%})} \\
SparseVLM (ICML2025)& 51.2 & 63.1 & 1542 & 77.3 & 67.5 & 66.3 & 46.4 & - & 85.0\% \\
VisionZip (CVPR2025)& 55.5 & 60.1 & 1630 & 74.8 & 68.3 & 71.4 & \textbf{56.2} & 55.5 & 90.3\% \\
DivPrune (CVPR2025)& 57.8 & 62.0 & 1658 & 79.4 & 68.0 & 73.9 & 52.4 & \textbf{56.1} & 91.6\% \\
Zoo-Prune (CVPR2026)& 59.9 & \textbf{64.2} & 1738 & 83.1 & \textbf{68.4} & 76.1 & 55.4 & - & 93.5\% \\
\rowcolor{orange!15} STS (Ours) & \textbf{61.1} & 63.7 & \textbf{1765} & \textbf{87.1} & 67.8 & \textbf{76.9} & 55.4 & 55.8 & \textbf{95.3\%} \\
\bottomrule
\end{tabular}
}
\caption{Performance comparison of different token pruning methods on LLaVA-NeXT-7B across multiple benchmarks. The orange background highlights our method.}
\label{tab:next}
\end{table*}

\begin{table*}[t]
\centering

\resizebox{\textwidth}{!}{
\begin{tabular}{l|cccccc|c}
\toprule
\textbf{Method} & \textbf{\# Token} & \textbf{FLOPs (T)} & \textbf{Prefill (ms)} & \textbf{Decode (ms)} & \textbf{KV Cache (MB)} & \textbf{Memory (GB)} & \textbf{Score (F1)} \\
\midrule
\rowcolor{gray!15} \multicolumn{8}{c}{\textit{Upper Bound, All 2880 Tokens (100\%)}} \\
LLaVA-NeXT-7B & 2880 & 41.7 & 246 & 29 & 1440.0 & 16.7 & 86.8 \\
\midrule
\rowcolor{gray!15} \multicolumn{8}{c}{\textit{Retain 320 Tokens ($\downarrow$ 88.9\%)}} \\
FastV (ECCV24) & 320 & 4.4 ($\times$9.5) & 54 ($\times$4.6) & 23 ($\times$1.2) & 160.3 & 15.6 & 49.5 \\
SparseVLM (ICML25) & 320 & 4.5 ($\times$9.3) & 71 ($\times$3.5) & 25 ($\times$1.1) & 161.2 & 18.6 & 76.9 \\
VisionZip (CVPR25) & 320 & \textbf{4.2} ($\times$9.9) & \textbf{38} ($\times$6.6) & \textbf{22} ($\times$1.3) & \textbf{160.0} & 14.8 & 82.3 \\
DivPrune (CVPR25) & 320 & \textbf{4.2} ($\times$9.9) & \textbf{38} ($\times$6.6) & \textbf{22} ($\times$1.3) & \textbf{160.0} & \textbf{13.8} & 84.7 \\
\rowcolor{orange!15} STS (Ours) & 320 & \textbf{4.2} ($\times$9.9) & \textbf{38} ($\times$6.6) & \textbf{22} ($\times$1.3) & \textbf{160.0} & \textbf{13.8} & \textbf{87.7} \\
\bottomrule
\end{tabular}
}

\caption{Efficiency Analysis on LLaVA-NeXT-7B. The orange background highlights our method.}
\label{tab:efficiency_analysis}

\vspace{1.2em}

\includegraphics[width=\textwidth]{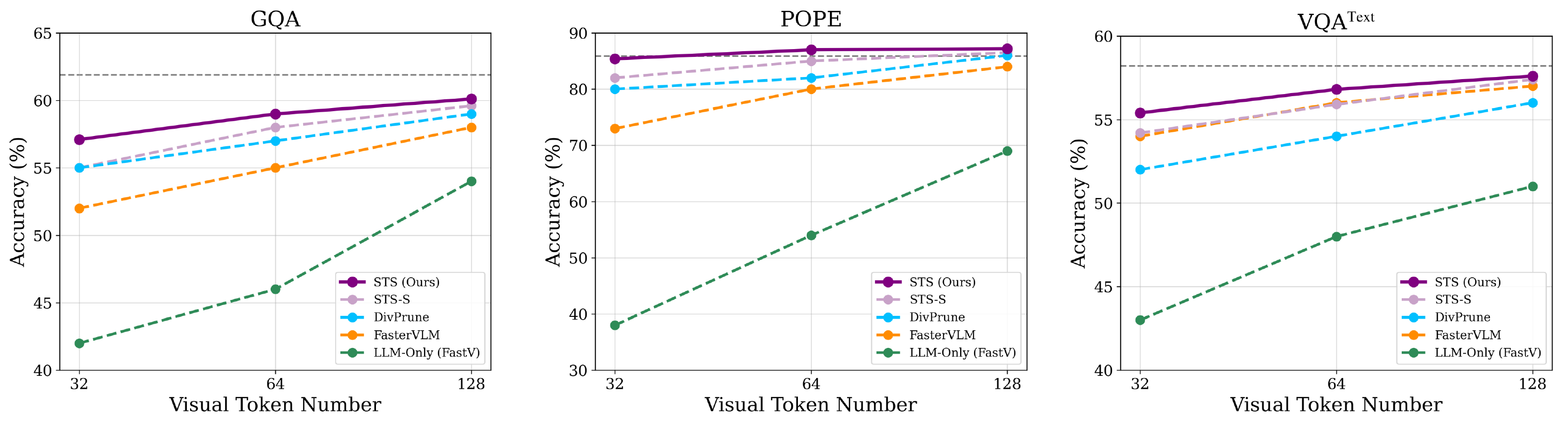}

\captionof{figure}{\textbf{Performance comparison of different token pruning variants under rigorous budgets.}
Results are reported on GQA, POPE, and TextVQA using LLaVA-NeXT-7B. The full STS framework consistently outperforms all single-stage or attention-dependent baselines across varying preserved token counts.}
\label{fig:xr}

\end{table*}

\subsection{Ablation Studies }

To analyze the effectiveness of our decoupled design, we compare the full STS framework with four representative variants on LLaVA-1.5-7B: \textbf{LLM-Only (FastV)}, which performs pruning using LLM attention; \textbf{FasterVLM}, which uses standard \texttt{[CLS]} attention after the vision encoder; \textbf{DivPrune}, which selects tokens through max-min diversity; and \textbf{STS-S}, our Stage-1-only variant that uses the potential-energy objective without task-aware semantic filtering. We report results on GQA, POPE, and TextVQA under different pruning budgets to examine the roles of pre-LLM diversity preservation and intra-LLM semantic filtering.

First, attention-based pruning methods degrade substantially under aggressive token reduction. FasterVLM, which prunes visual tokens at the ViT exit based on attention scores, drops to 73.0\% on POPE at the lowest budget, compared with the 85.9\% full-token baseline. FastV performs even worse, falling to 38.0\% on POPE and 42.0\% on GQA. As shown in our visualization, attention-only pruning can also concentrate tokens in prompt-irrelevant regions, such as the lower-right area of the image, wasting the limited token budget . As showed in \ref{C}. In contrast, STS first constructs a diverse candidate set before LLM-stage filtering, helping the final selection focus more on useful visual content.

Second, among pre-LLM diversity-preserving methods, STS-S consistently outperforms DivPrune. At the most aggressive pruning ratio, STS-S achieves 82.0\% on POPE and 54.0\% on TextVQA, compared with 81.0\% and 52.0\% for DivPrune. Since both methods operate without textual instructions, this suggests that the potential-energy objective provides a stronger structural selection criterion than standard distance-based diversity selection by considering each candidate token relative to the entire selected set.

Third, intra-LLM task-aware filtering further improves over STS-S. At the lowest budget, full STS improves POPE from 82.0\% to 85.4\%, showing that structural diversity should be refined according to the textual query. At higher budgets, STS reaches 87.2\% on POPE, slightly exceeding the unpruned baseline of 85.9\%. Overall, these results indicate that the two stages are complementary: the first preserves a diverse candidate set, while the second removes prompt-irrelevant tokens.

\subsection{Efficiency Analysis  }

Table \ref{tab:efficiency_analysis} reports the efficiency--performance trade-off on LLaVA-NeXT-7B under an aggressive 88.9\% token reduction. STS reduces FLOPs by 9.9$\times$ and prefill latency by 6.6$\times$ (from 246 ms to 38 ms). It also reduces the KV cache from 1440.0 MB to 160.0 MB and lowers peak memory usage to 13.8 GB. Compared with SparseVLM, which introduces additional memory and latency overhead, STS matches the efficiency profile of lightweight baselines such as VisionZip and DivPrune.

Despite this substantial reduction in computation, STS achieves an F1 score of 87.7, slightly exceeding the 2880-token unpruned baseline (86.8) under the same setting. In contrast, FastV and SparseVLM exhibit much larger performance drops at this compression ratio. These results indicate that STS provides a strong efficiency--performance trade-off in high-redundancy settings.

\section{Conclusion}

We presented STS, a training-free, stage-aware visual token pruning framework for efficient multimodal inference. Our analysis shows that attention-based pruning tends to select high-attention tokens from semantically similar regions, causing redundant selection and reduced diversity. By combining diversity-preserving pre-LLM selection with task-aware intra-LLM filtering, STS improves efficiency under aggressive token reduction across VLM architectures while maintaining task performance. These results highlight the importance of matching pruning strategies to representation dynamics.

\section*{Limitations}

Although the STS framework preserves strong task performance under aggressive token reduction without requiring additional training, it is not entirely without loss. Furthermore, like many efficiency-oriented approaches that operate on intermediate model representations, STS requires direct access to internal visual tokens during inference. As a result, it cannot be directly applied to black-box multimodal models, such as proprietary GPT- or Claude-style systems, where such intermediate representations are not exposed. Moving forward, we are committed to advancing our research toward model quantization and broader efficient LLM paradigms, aiming to develop more versatile and lossless methods to further enhance the efficiency of visual understanding.

\bibliography{main}

\clearpage 
\appendix  

\section*{Appendix }

\section{Additional Analysis and Algorithm Details}
\label{A}
\subsection{Background and Rationale   }

K-nearest neighbors (KNN) is used in this work not as a classifier, but as a diagnostic tool for probing the local geometry of visual token representations. Specifically, we use KNN to examine whether visual tokens that are close in the feature space also receive similar attention-based importance scores. This question is central to understanding the failure mode of attention-based pruning: if nearby tokens receive nearly identical attention scores, attention becomes less discriminative within local neighborhoods and may lead to redundant token selection.

Given a set of $N$ visual tokens at a certain layer, each token $t_i$ is represented by a feature vector $\mathbf{f}_i \in \mathbb{R}^d$ and an attention score $s_i \in \mathbb{R}$. For each token, we construct a local neighborhood $\mathcal{N}_K(t_i)$ using the $K$ nearest neighbors in the feature space. In our implementation, cosine distance is used to reduce the influence of feature magnitude in high-dimensional representations.

To measure how attention scores vary within these local neighborhoods, we define the local fluctuation as
\begin{equation}
    \sigma_{\text{local}} = 
    \frac{1}{N} \sum_{i=1}^{N} 
    \operatorname{std}\left(
    \{ s_j \mid t_j \in \mathcal{N}_K(t_i) \}
    \right).
\end{equation}
We further normalize this quantity by the global attention fluctuation
\begin{equation}
    \sigma_{\text{global}} =
    \operatorname{std}(\{s_i\}_{i=1}^{N}),
\end{equation}
and define the KNN Consistency Score as
\begin{equation}
    C = 1 - \frac{\sigma_{\text{local}}}{\sigma_{\text{global}}}.
\end{equation}
A low value of $C$ indicates that attention scores vary substantially even among feature-similar tokens, suggesting weak alignment between attention similarity and feature similarity. In contrast, a high value of $C$ indicates that tokens within the same local feature neighborhood tend to receive similar attention scores. This implies that attention becomes locally homogeneous in the feature space, making attention-based pruning more likely to retain or discard groups of semantically similar tokens together.

This analysis helps explain the Manifold Coverage Gap observed in attention-based pruning. When attention scores become locally consistent within feature neighborhoods, pruning based only on attention magnitude may select redundant tokens from the same semantic region while missing complementary long-tail visual details. Therefore, the KNN Consistency Score provides empirical evidence that attention magnitude alone is insufficient for preserving diverse visual representations in deep vision layers.

KNN is suitable for this analysis because it captures local neighborhood structure without imposing hard cluster assignments. Compared with clustering methods, which require discrete partitioning of the feature space, KNN provides a more flexible way to probe local feature geometry. Compared with spectral methods, which emphasize global structure, KNN directly focuses on local behavior, making it well suited for detecting redundancy among semantically similar visual tokens.

\subsection{Cross-Model Consistency Analysis   }

To assess whether this phenomenon is consistent across different vision encoders, we extend our KNN-based analysis to a broader set of representative architectures. We focus on whether, as encoder depth increases, tokens with similar attention scores tend to become more concentrated in nearby regions of the feature space. Specifically, we evaluate the following vision encoders:

\textbf{LLaVA-1.5 } \citep{liu2024improvedbaselinesvisualinstruction}: As a cornerstone of open-source MLLMs, LLaVA-1.5 utilizes a pretrained CLIP visual encoder connected to a Vicuna language model via an MLP projector. This architecture processes 336×336 resolution images, resulting in 576 visual tokens, and achieves state-of-the-art performance through extensive visual instruction tuning. 

\textbf{InternVL3 } \citep{zhu2025internvl3exploringadvancedtraining} :This model adopts a native multimodal pre-training paradigm within a ViT-MLP-LLM framework, acquiring linguistic and multimodal capabilities simultaneously. By incorporating Variable Visual Position Encoding, InternVL3 demonstrates superior performance in handling extended contexts and specialized tasks such as industrial image analysis and 3D perception. 

\textbf{Qwen2.5-VL} \citep{bai2023qwentechnicalreport}:  Representing the latest advancement in the Qwen-VL series, this model employs a redesigned Vision Transformer architecture featuring window attention and dynamic resolution support. It excels in complex visual reasoning, document parsing, and long-video comprehension, functioning as a versatile agent capable of precise event localization and tool usage. 

\begin{figure*}[t] 
    \centering
    \includegraphics[width=\textwidth]{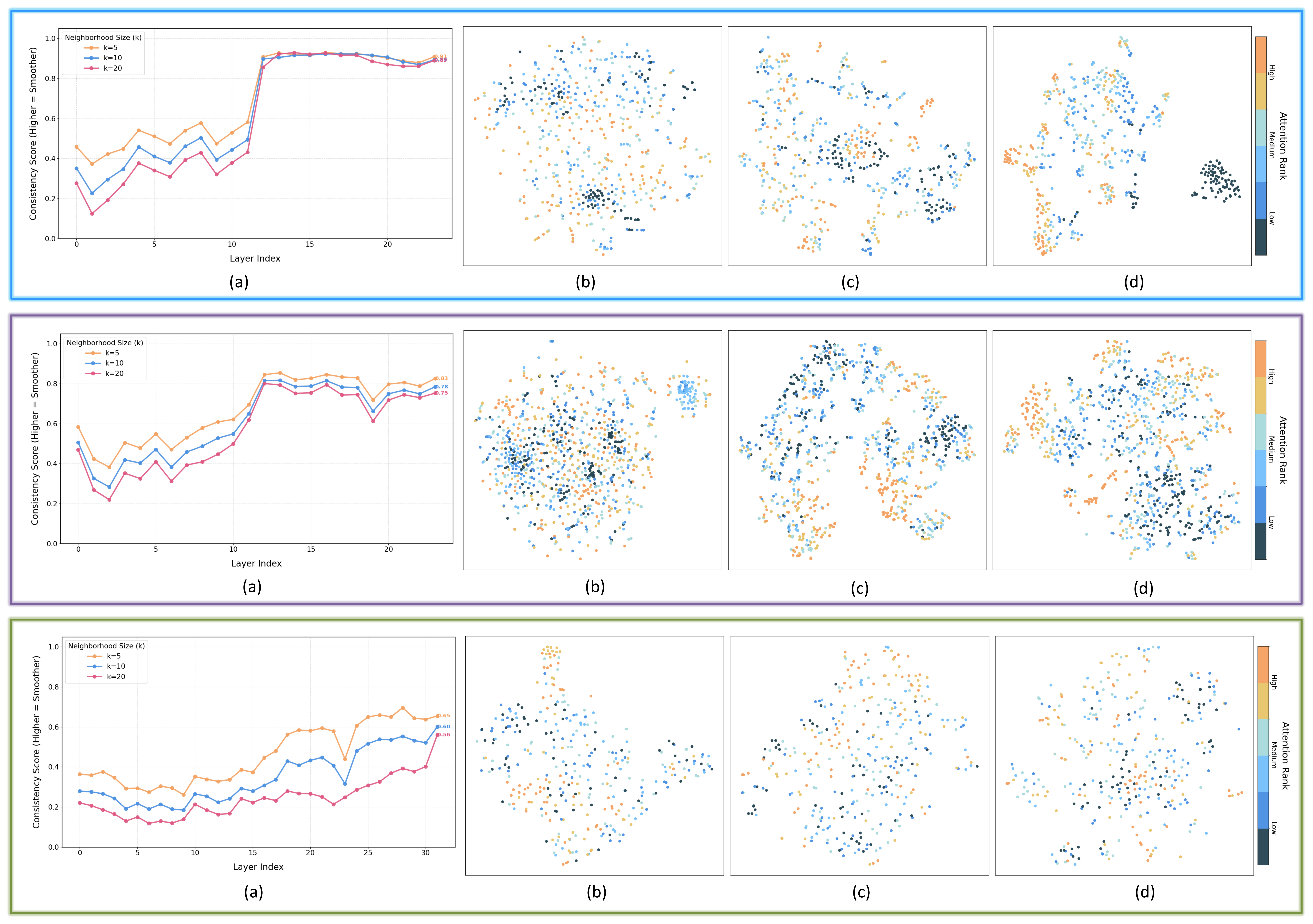} 
    
   \caption{Evolution of feature redundancy across diverse VLM vision encoders. The figure compares LLaVA-1.5 (blue box), InternVL3 (purple box), and Qwen2.5-VL (green box). Panel (a) displays the KNN Consistency Score across layers for different $k$ values, consistently revealing a sharp increase in deeper layers. Panels (b)-(d) show t-SNE visualizations of token embeddings at shallow, middle, and deep layers (Layers 1, 12, 24 for LLaVA/InternVL3; Layers 1, 16, 32 for Qwen2.5-VL), colored by their attention ranks. The cross-model visual evidence clearly demonstrates a universal dispersion-to-aggregation pattern: tokens in shallow layers are broadly distributed, whereas high-attention tokens in deep layers collapse into redundant, localized clusters.}
    \label{fig:knn_3}
\end{figure*}

\subsection{Observations and Visualization    }

To provide a cross-model comparison, we visualize both the quantitative KNN sensitivity analysis and the corresponding qualitative feature distributions. Figure~\ref{fig:knn_3} presents the results for LLaVA-1.5, InternVL3, and Qwen2.5-VL.

\paragraph{Quantitative Trend: Layer-wise Increase in Local Consistency.}
As shown in Panel (a), all three models exhibit a similar layer-wise trend under different neighborhood sizes ($k \in \{5, 10, 20\}$). In shallow and intermediate layers, the KNN Consistency Scores remain relatively low, suggesting weak alignment between feature similarity and attention similarity. In other words, tokens that are close in the feature space do not necessarily receive similar attention scores at these layers. As depth increases, the consistency scores gradually rise across different values of $k$, indicating that attention scores become more locally consistent within feature neighborhoods. This trend suggests that deeper vision layers increasingly assign similar attention scores to feature-similar tokens, which may increase redundancy for attention-based token selection.

\paragraph{Qualitative Evidence: From Dispersion to Concentration.}
Panels (b), (c), and (d) show t-SNE projections of visual token embeddings at shallow, intermediate, and deep layers, respectively. Points are colored by attention rank, with warmer colors indicating higher attention scores. In shallow layers, token embeddings are relatively dispersed in the projected space, and high-attention tokens are distributed across different regions. In deeper layers, tokens with similar attention ranks become more concentrated in nearby regions, indicating stronger alignment between attention patterns and local feature neighborhoods.

Overall, the quantitative and qualitative results reveal a consistent depth-dependent pattern across the evaluated vision encoders: attention scores become increasingly aligned with local feature similarity in deeper layers. This observation supports our motivation for introducing a diversity-preserving pre-LLM pruning stage, which aims to reduce redundant token selection before task-aware filtering in the LLM.

\begin{algorithm}[t]
\caption{STS: Structure-to-Semantics Visual Token Pruning}
\label{alg:sts}
\begin{algorithmic}[1]
\REQUIRE Image $I$, prompt $P$, pre-LLM budget $K$, pruning layers $\mathcal{L}$, intra-LLM budgets $\mathcal{B}$
\ENSURE Output logits

\STATE $\mathbf{V} \leftarrow \mathrm{VisionTower}(I)$
\STATE $\mathcal{S}\leftarrow \emptyset$, $r_i \leftarrow 0$ for all visual tokens

\FOR{$t=1$ to $K$}
    \STATE $i^\ast \leftarrow \arg\min_{i\notin \mathcal{S}} r_i$
    \STATE $\mathcal{S}\leftarrow \mathcal{S}\cup\{i^\ast\}$
    \STATE $r_i \leftarrow r_i + \frac{1}{\|\mathbf{v}_i-\mathbf{v}_{i^\ast}\|_2^2+\epsilon}$ for all $i$
\ENDFOR

\STATE $\mathbf{V}_{\mathcal{S}} \leftarrow \mathrm{MMProjector}(\mathbf{V}_{\mathcal{S}})$
\STATE Construct multimodal embeddings $\mathbf{H}$ from $P$ and $\mathbf{V}_{\mathcal{S}}$

\FOR{each LLM layer $\ell$}
    \STATE $\mathbf{H}\leftarrow \mathrm{DecoderLayer}_{\ell}(\mathbf{H})$
    \IF{$\ell\in\mathcal{L}$ and in prefill stage}
        \STATE Compute attention from the last instruction token to visual tokens
        \STATE Select top-ranked visual tokens according to attention scores
        \STATE Rebuild $\mathbf{H}$, attention mask, and position IDs
    \ENDIF
\ENDFOR

\RETURN $\mathrm{LMHead}(\mathrm{Norm}(\mathbf{H}))$
\end{algorithmic}
\end{algorithm}

\section{Supplementary Experiments  }
\label{B}

In this section, we provide extended experimental results to demonstrate the scalability and architectural generalizability of our proposed STS framework. Specifically, we evaluate STS on a larger model scale (LLaVA-1.5-13B) and a fundamentally different VLM architecture (Qwen2.5-VL-7B). Furthermore, we provide a detailed ablation study on the optimal layer depth for executing the Stage-2 semantic pruning within the LLM.

\subsection{Evaluation benchmarks}

\textbf{GQA.} \citep{hudson2019gqanewdatasetrealworld} GQA integrates three key components: scene graphs, questions, and images, providing both spatial and object-level visual features. The questions are specifically structured to rigorously test a model's capacity for visual scene reasoning and compositional understanding.

\textbf{MMBench.} \citep{li2023evaluatingobjecthallucinationlarge} Designed for a holistic assessment, MMBench employs a hierarchical evaluation framework spanning three granularity levels (L-1 to L-3). It begins with broad capabilities like perception and reasoning and refines them into 20 specific dimensions. This multi-tiered structure allows for a fine-grained analysis of the model's comprehensive abilities.

\textbf{MME.} \citep{fu2025mmecomprehensiveevaluationbenchmark} MME is a comprehensive suite targeting both perceptual and cognitive competencies through 14 distinct subtasks. By leveraging concise instruction designs and manually curated instruction-answer pairs, it effectively minimizes the risks of data leakage and ensures a fair, robust evaluation of model performance.

\textbf{POPE.} \citep{pope2023efficiently} Focusing on object hallucination, POPE reformulates evaluation as a series of binary queries regarding object presence. It utilizes robust metrics—including Accuracy, Precision, Recall, and F1 Score—to strictly quantify hallucination rates across three distinct sampling settings.

\textbf{ScienceQA.} \citep{lu2022learnexplainmultimodalreasoning} Spanning natural, language, and social sciences, ScienceQA features a rich taxonomy structured by topic, category, and skill (covering 26, 127, and 379 distinct types, respectively). This benchmark serves as a rigorous testbed for multimodal understanding, multi-hop reasoning, and interpretability within scientific contexts.

\textbf{VQA-v2.} \citep{goyal2017makingvvqamatter} This benchmark assesses visual perception via open-ended questioning on a massive scale. Comprising over 265,000 images representing diverse real-world scenarios, VQA-v2 utilizes 10 human-annotated ground truth answers per question to ensure accurate and reliable performance benchmarking.

\textbf{TextVQA.} \citep{singh2019vqamodelsread} TextVQA targets the interpretation of textual information embedded within visual scenes. It demands that models not only perceive visual content but also detect, read, and reason about text in images to answer questions accurately, thereby evaluating integrated optical character recognition (OCR) and reasoning skills.

\textbf{VizWiz.}  \citep{gurari2018vizwizgrandchallengeanswering}A visual question answering benchmark collected in a real-world accessibility setting, where blind users captured images and asked spoken questions about them. Each visual question is paired with 10 crowdsourced answers. It introduces two key tasks: answering visual questions and predicting whether a question is unanswerable based on the image, highlighting challenges such as poor image quality and ambiguous content. We use the test split for evaluation. 

\begin{table*}[t!]
\centering
\resizebox{\textwidth}{!}{
\begin{tabular}{l|cccccccc|c}
\toprule
\textbf{Method} & \textbf{GQA} & \textbf{MMB} & \textbf{MME} & \textbf{POPE} & \textbf{SQA} & \textbf{VQA$^{\text{v2}}$} & \textbf{VQA$^{\text{Text}}$} & \textbf{VizWiz} & \textbf{Avg.} \\
\midrule
\rowcolor{gray!15} LLaVA-1.5-13B & \multicolumn{9}{c}{\textit{Upper Bound (100\%)}} \\
Vanilla & 63.2 & 67.7 & 1818 & 85.9 & 72.8 & 80.0 & 61.3 & 53.6 & 100.0\% \\
\midrule
\rowcolor{gray!15} LLaVA-1.5-13B & \multicolumn{9}{c}{\textit{Retain 128 Tokens}} \\
VisionZip & 57.9 & 66.7 & 1743 & 85.2 & 74.0 & 76.8 & 58.9 & 52.3 & 96.5\% \\
DART & 57.7 & 65.4 & 1751& 80.4 & 74.2 & 75.7 & 58.7 & 53.0 & 96.8\% \\
DivPrune & 58.9 & 66.1 & 1748 & 86.5 & 72.8 & 77.1 & 58.2 & 53.5 & 97.6\% \\
AgilePrune & 59.1 & 67.6 & - & 86.9 & 72.8 & 77.5 & 58.9 & 52.5 & 98.4\% \\
Zoo-Prune & 58.9 & 67.0 & 1791 & 87.0 & 73.4 & 77.8 & 58.8 & - & 97.6\% \\
\rowcolor{orange!15} STS (Ours) & \textbf{60.1} & \textbf{67.8} & \textbf{1800} & \textbf{87.9} & 73.2 & \textbf{78.5} & 58.9 & 52.8 & \textbf{99.4\%} \\
\midrule
\rowcolor{gray!15} LLaVA-1.5-13B & \multicolumn{9}{c}{\textit{Retain 64 Tokens}} \\
VisionZip & 56.2 & 64.9 & 1676 & 76.0 & 74.4 & 73.7 & 57.4 & 53.2 & 94.1\% \\
DART & 55.7 & 64.7 & 1769& 72.8 & 73.8 & 72.4 & 57.3 & 53.4 & 93.9\% \\
DivPrune & 57.7 & 64.6 & 1778 & 84.8 & 71.3 & 75.2 & 57.1 & 54.4 & 94.8\% \\
AgilePrune & 57.5 & 66.2 & - & 82.0 & 72.0 & 75.7 & 58.6 & 54.2 & 96.9\% \\
Zoo-Prune & 58.6 & 64.8 & 1780 & 85.3 & 72.1 & 76.4 & 58.6 & - & 95.2\% \\
\rowcolor{orange!15} STS (Ours) & \textbf{58.7} & \textbf{66.3} & \textbf{1789} & \textbf{87.2} & 72.6 & \textbf{77.2} & \textbf{58.9} & 52.8 & \textbf{98.0\%} \\
\midrule
\rowcolor{gray!15} LLaVA-1.5-13B & \multicolumn{9}{c}{\textit{Retain 32 Tokens}} \\
VisionZip & 52.7 & 61.2 & - & 66.8 & 72.9 & 68.4 & 55.2 & 53.0 & 89.8\% \\
DART & 53.9 & 61.9 & - & 66.9 & 73.2 & 68.1 & 55.1 & 52.0 & 90.9\% \\
DivPrune & 56.2 & 61.7 & - & 79.3 & 70.9 & 72.0 & 54.6 & 54.5 & 93.1\% \\
\rowcolor{orange!15} STS (Ours) & \textbf{57.9} & \textbf{64.1} & \textbf{1734} & \textbf{86.5} & \textbf{72.6} & \textbf{75.4} & \textbf{57.5} & 53.9 & \textbf{96.0\%} \\
\bottomrule
\end{tabular}
}
\caption{Performance comparison of different token pruning methods on LLaVA-1.5-13B across multiple benchmarks. The orange background highlights our proposed STS method.}
\label{tab:13b}
\end{table*}
\begin{table}[t!]
\centering
\resizebox{\columnwidth}{!}{
\begin{tabular}{l|cccc|c}
\toprule
\textbf{Method} & \textbf{GQA} & \textbf{MMB} & \textbf{MME} & \textbf{POPE} & \textbf{Rel.} \\
\midrule
\rowcolor{gray!15} \multicolumn{6}{c}{\textit{Baseline (Full Tokens)}} \\
Qwen2.5-VL-7B & 60.84 & 84.10 & 2310 & 86.30 & 100.0\% \\
\midrule
\rowcolor{gray!15} \multicolumn{6}{c}{\textit{Retain 20\% Tokens}} \\
VisionZip  & 57.27 & 79.72 & \textbf{2221} & 83.89 & 95.6\% \\
DivPrune  & 60.05 & 79.55 & 2173 & 83.42 & 96.0\% \\
\rowcolor{orange!15} STS (Ours) & \textbf{60.31} & \textbf{80.70} & 2211 & \textbf{84.37} & \textbf{97.1\%} \\
\midrule
\rowcolor{gray!15} \multicolumn{6}{c}{\textit{Retain 10\% Tokens}} \\
VisionZip  & 54.09 & 76.03 & 1937 & 78.97 & 88.7\% \\
DivPrune  & 55.49 & 76.03 & 2054 & 79.05 & 90.5\% \\
\rowcolor{orange!15} STS (Ours) & \textbf{56.35} & \textbf{76.48} & \textbf{2108} & \textbf{80.97} & \textbf{92.2\%} \\
\bottomrule
\end{tabular}
}
\caption{Performance comparison on Qwen2.5-VL-7B under different token retention ratios. The orange background highlights our proposed STS method.}
\label{tab:qwen}
\end{table}
\begin{table*}[t!]
\centering
\begin{tabular}{c|ccccc}
\toprule
\textbf{Settings} & \textbf{GQA} & \textbf{VQA$^{\text{text}}$} & \textbf{MME} & \textbf{SQA} & \textbf{POPE} \\
\midrule
$L_{\text{prune}} = 2, \rho_{\text{intra}} = 73.3\%$ & 60.2 & 57.6 & 1805 & \textbf{69.7\%} & 86.2 \\
$L_{\text{prune}} = 8, \rho_{\text{intra}} = 66.7\%$ & 60.1 & 57.6 & 1814 & 69.3\% & 85.9 \\
$L_{\text{prune}} = 12, \rho_{\text{intra}} = 60.0\%$ & \textbf{60.9} & 57.8 & \textbf{1816} & 69.6\% & \textbf{86.3} \\
$L_{\text{prune}} = 16, \rho_{\text{intra}} = 50.0\%$ & \textbf{60.9} & \textbf{58.0} & 1793 & 69.6\% & \textbf{86.3} \\
$L_{\text{prune}} = 20, \rho_{\text{intra}} = 33.3\%$ & 60.8 & 57.8 & 1803 & 69.6\% & 86.2 \\
$L_{\text{prune}} = 24, \rho_{\text{intra}} = 0.0\%$  & 60.6 & 57.4 & 1793 & 69.4\% & 86.2 \\
\bottomrule
\end{tabular}
\caption{Ablation study on hyper-parameters $L_{\text{prune}}$ and $\rho_{\text{intra}}$.}
\label{tab:ablation_settings}
\end{table*}

\subsection{Generalization to Larger Scales and Dynamic Architectures}

\textbf{Results on LLaVA-1.5-13B.} As shown in Table \ref{tab:13b}, STS also performs consistently well on the larger LLaVA-1.5-13B model. With 128 retained tokens, STS reaches 99.4\% relative performance, outperforming recent baselines such as AgilePrune and Zoo-Prune. When the budget is reduced to 64 tokens, STS retains 98.0\% relative performance and achieves the best results on GQA, MMB, MME, POPE, VQA$^{\text{v2}}$, and VQA$^{\text{Text}}$. Even under the extreme 32-token setting, STS maintains 96.0\% relative performance, exceeding DivPrune by 2.9 points and showing that the proposed pruning strategy remains effective on larger VLMs.

\textbf{Results on Qwen2.5-VL-7B.}
We further evaluate STS on Qwen2.5-VL-7B to examine its effectiveness beyond the LLaVA family. As shown in Table \ref{tab:qwen}, STS achieves 97.1\% relative performance when retaining 20\% of visual tokens, outperforming VisionZip and DivPrune by 1.5 and 1.1 points, respectively. Under the more aggressive 10\% token budget, STS retains 92.2\% relative performance and achieves the best results across all evaluated benchmarks. These results indicate that STS remains effective on dynamic-resolution VLM architectures.

\subsection{ Ablation on the pruning layer.}

We analyze the sensitivity of STS to the intra-LLM pruning layer $L_{\text{prune}}$. In this experiment, we fix the pre-LLM retention ratio as $\rho_{\text{pre}}=50\%$ and keep the overall retention ratio at 37.5\% by adjusting $\rho_{\text{intra}}$ for different choices of $L_{\text{prune}}$.

As shown in Table \ref{tab:ablation_settings}, the performance remains relatively stable across a wide range of pruning layers. For example, GQA varies only from 60.1 to 60.9, VQA$^{\text{text}}$ from 57.4 to 58.0, and POPE from 85.9 to 86.3, indicating that STS is not highly sensitive to the exact choice of $L_{\text{prune}}$. Meanwhile, intermediate layers such as $L_{\text{prune}}=12$ and $L_{\text{prune}}=16$ achieve the best overall results, suggesting that middle layers provide a suitable pruning point where task-relevant signals have emerged while redundant visual tokens can still be removed before later computation.

\section{Additional Visualization Results}
\label{C}

\subsection{Visualization of Pre-LLM Token Selection }

\begin{figure*}[t] 
    \centering
    \includegraphics[width=\textwidth]{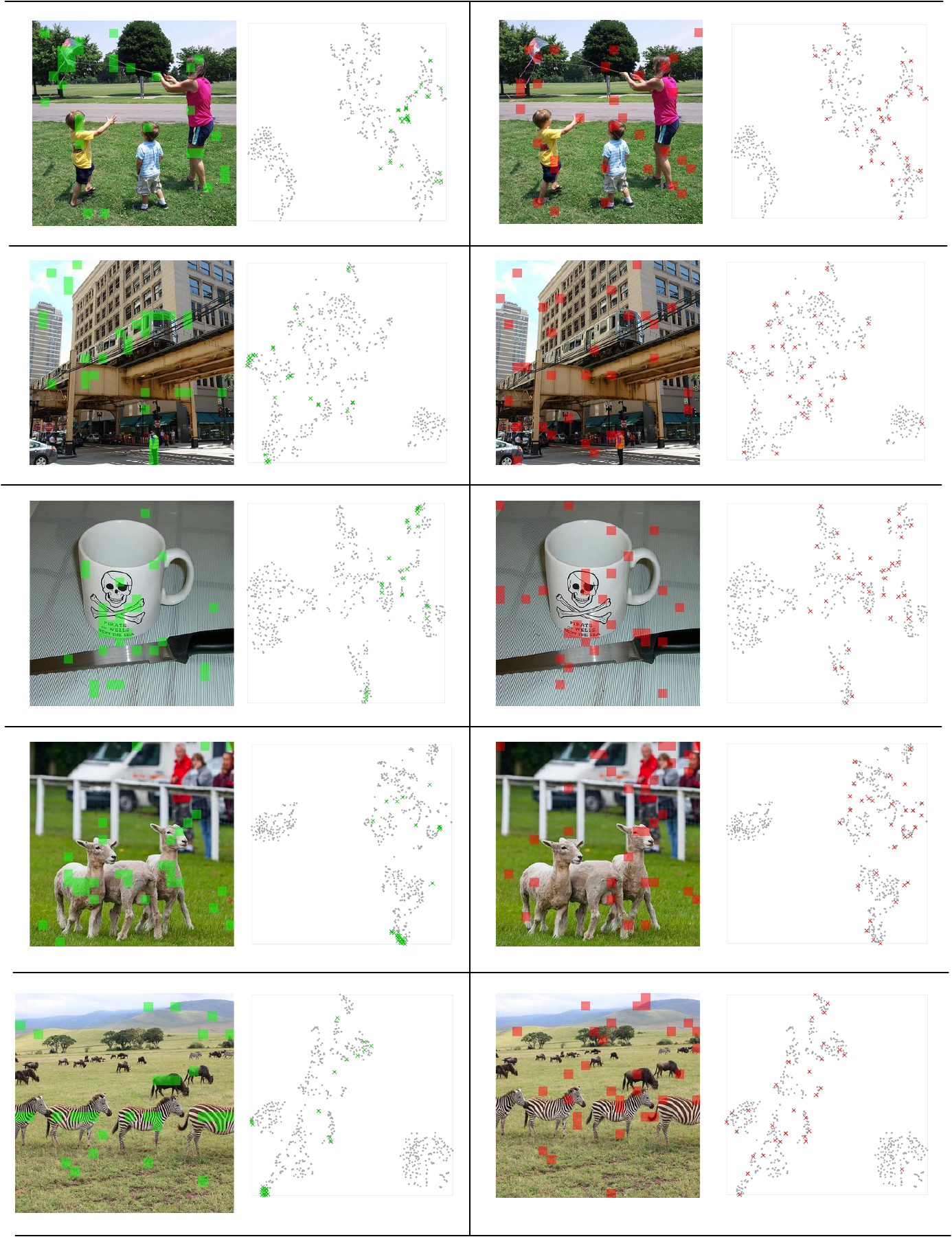} 
    
  \caption{
Visualization of vision-encoder token selection under a 32-token budget.
Green markers denote tokens selected by the \texttt{[CLS]} attention-based method,
and red markers denote tokens selected by STS-S.
}
    \label{fig:vit_tsne}
\end{figure*}

We first present visualizations of vision-encoder token selection using image masks and t-SNE projections. Under the same 32-token budget, we compare STS-S with \texttt{[CLS]} attention-based selection in Figure \ref{fig:vit_tsne}. The attention-based strategy tends to retain tokens from a few high-response regions, which often results in redundant selections of semantically similar visual content. As a result, several visually distinct regions in the image and feature space may receive limited coverage.

By contrast, STS-S produces a more balanced and informative token subset. Since the potential-energy objective penalizes selecting tokens that are close to already retained ones in the feature space, the selected tokens naturally spread toward different semantic regions. Interestingly, this process does not simply scatter tokens uniformly at random; instead, it tends to retain tokens from visually diverse and information-rich areas, including objects, contextual regions, boundaries, and other less dominant but complementary visual cues. In the image masks, STS-S covers a broader range of meaningful regions, while in the t-SNE space, the retained tokens span multiple feature clusters rather than collapsing into a single dense area. This suggests that potential-energy-based selection can automatically construct a less redundant visual summary before the tokens are passed to the LLM.

\begin{figure*}[t] 
    \centering
    \includegraphics[width=\textwidth]{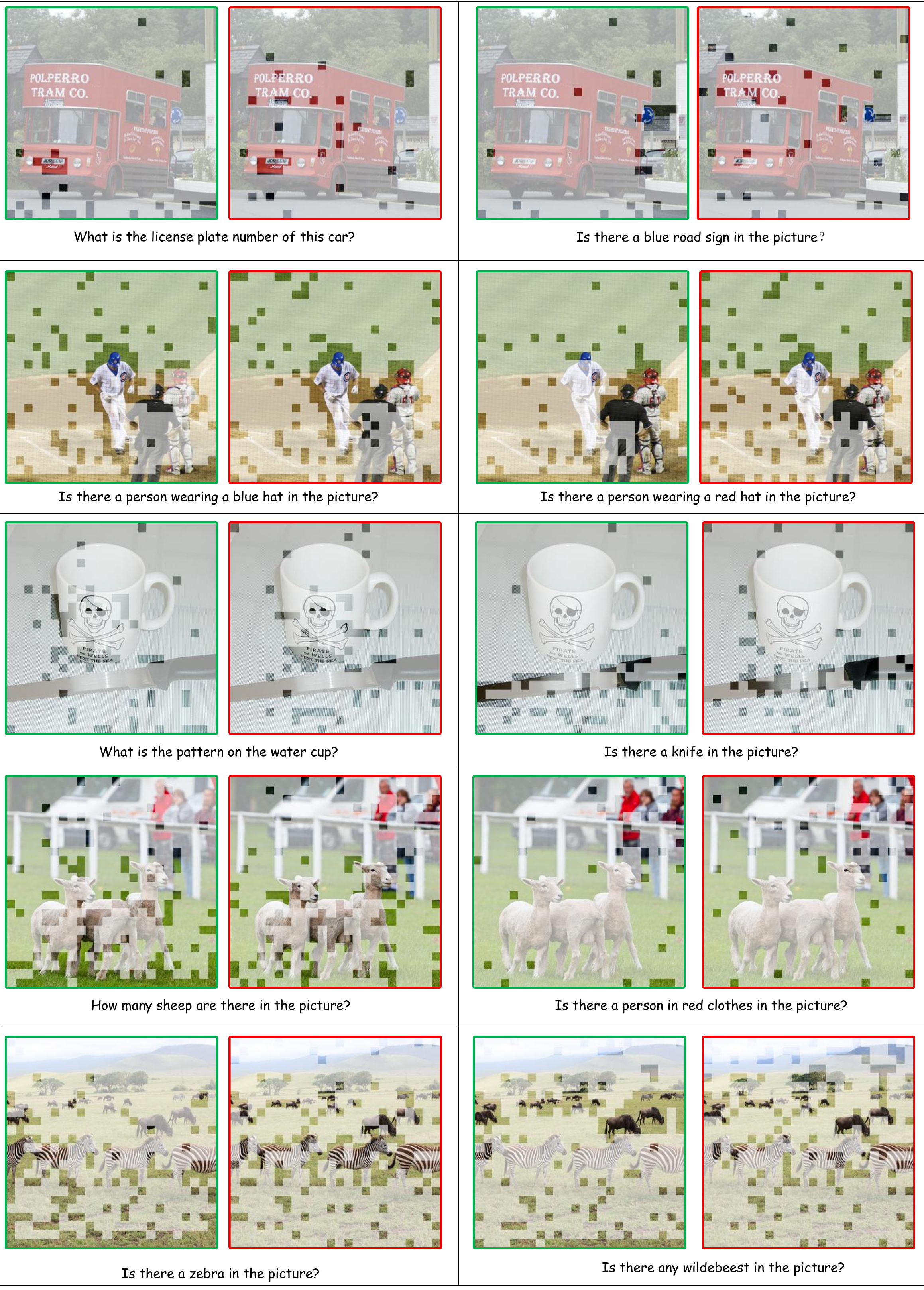} 
    
\caption{
Visualization of final token selection.
Green boxes denote attention-based pruning results, and red boxes denote STS results.
}
    \label{fig:mark}
\end{figure*}

\subsection{Visualization of the Complete STS Process  }

In this visualization, we compare the final token selections of STS with an LLM-only attention-based pruning strategy  in Figure  \ref{fig:mark}. We observe that when pruning relies solely on LLM attention scores, the selected tokens tend to concentrate in the lower-right region of the image \citep{zhao2025mcallavamanhattancausalattention,zhang2024seeingclearlylayertwo}. These tokens are largely unrelated to the textual prompt, indicating a potential attention bias that wastes the limited token budget and can negatively affect the final prediction.

In contrast, STS avoids this issue by first applying diversity-preserving selection before the tokens enter the LLM. As a result, the subsequent task-aware filtering is performed over a more balanced candidate set, allowing the retained tokens to better focus on visually relevant regions for the given prompt.

\label{sec:appendix}

\end{document}